\documentclass[10pt,twocolumn,letterpaper]{article}
\usepackage{cvpr}
\usepackage{times}
\usepackage{epsfig}
\usepackage{graphicx}
\usepackage{caption}
\usepackage{amsmath}
\usepackage{amssymb}
\usepackage{xcolor}
\usepackage{subcaption}
\usepackage{booktabs}
\usepackage{array}
\usepackage[percent]{overpic}
\usepackage{rotating}
\usepackage{tabularx}
\definecolor{grey}{rgb}{0.8,0.8,0.8}
\definecolor{olive}{rgb}{0.,0.55,0.27}
\newif\ifdraft
\draftfalse
\ifdraft
\newcommand{\dcc}[1]{{\color{purple}\textbf{D:} #1}}
\newcommand{\rhc}[1]{{\color{orange}\textbf{RH:} #1}}
\newcommand{\ahc}[1]{{\color{olive}\textbf{A:} #1}}
\newcommand{\rgc}[1]{{\color{blue}\textbf{RG:} #1}}

\newcommand{\rh}[1]{{\color{orange}#1}}
\newcommand{\ah}[1]{{\color{olive}#1}}

\else
\newcommand{\dcc}[1]{}
\newcommand{\rhc}[1]{}
\newcommand{\rgc}[1]{}
\newcommand{\ahc}[1]{}

\newcommand{\rh}[1]{{\color{black}#1}}

\newcommand{\ah}[1]{{\color{black}#1}}

\fi

\newcommand{\al}[1]{#1 \emph{et al.}}
\newcolumntype{?}{!{\vrule width 1pt}}

\newcommand{\ourmethod}{PointGMM}
\newcommand{\hgm}{hGMM}
\newcommand{\ind}{j}

\newcommand{\depth}{d}
\newcommand{\pc}{X}
\newcommand{\pcr}{\ah{X}}

\newcommand{\argmax}[1]{\underset{#1}{\operatorname{arg}\,\operatorname{max}}\;}

\newcommand{\reals}{\mathbb{R}}
\newcommand{\inreals}[2]{#1 \in \reals^{#2}}
\newcommand{\smax}{\emph{softmax}}
\newcommand{\defeq}{\mathrel{\stackrel{\makebox[0pt]{\mbox{\normalfont\tiny def}}}{=}}}
\newcommand{\tr}{t}
\newcommand{\sh}{c}

\usepackage[pagebackref=true,breaklinks=true,letterpaper=true,colorlinks,bookmarks=false]{hyperref}

\cvprfinalcopy 

\def\cvprPaperID{8309} 

\begin{document}

\title{PointGMM: a Neural GMM Network for Point Clouds} 
\vspace{-0.1cm}
\author{Amir Hertz \qquad Rana Hanocka\qquad Raja Giryes\qquad Daniel Cohen-Or  \vspace*{0.2cm} \\
Tel Aviv University
\vspace*{-0.2cm}
}
\maketitle


\global\csname @topnum\endcsname 0
\global\csname @botnum\endcsname 0
\begin{figure}
    \includegraphics[trim={.85cm 0cm .85cm .85cm},clip,width=\columnwidth]{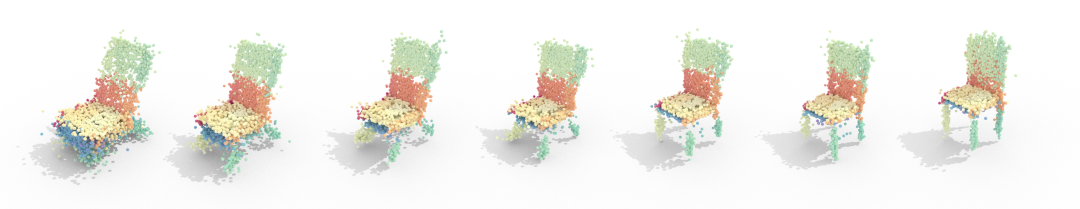}
    \\
    \includegraphics[trim={.85cm 0cm .85cm 2.3cm},clip,width=\columnwidth]{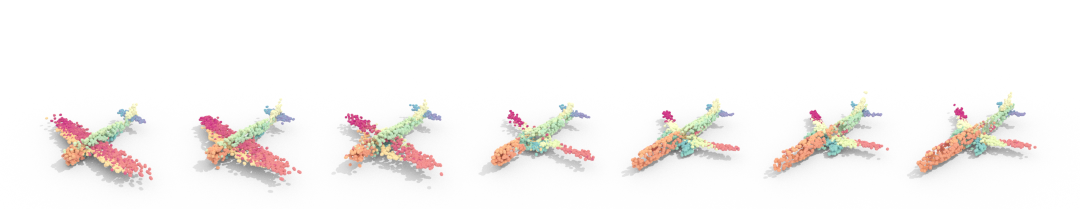}
    \\
    \includegraphics[trim={.85cm 0cm .85cm 1.5cm},clip,width=\columnwidth]{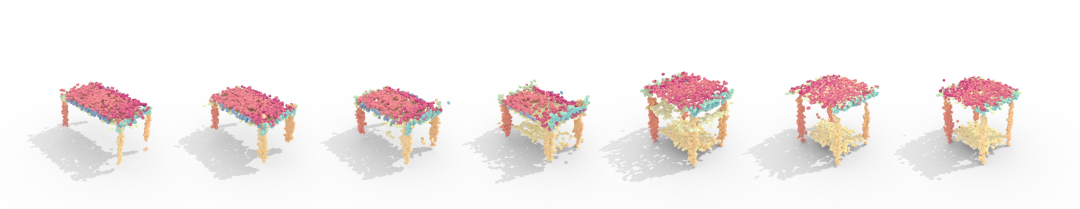}
    \caption{Shape interpolation using a \ourmethod{} generative model. The consistent coloring in the \hgm{} \rh{leaf} nodes suggests that an interpretable partitioning has been learned, without any supervision.}
    \label{fig:teaser}
    
\end{figure}

\begin{abstract}

Point clouds are a popular representation for 3D shapes. However, they encode a particular sampling without accounting for shape priors or non-local information.
We advocate for the use of a hierarchical Gaussian mixture model (\hgm{}), which is a compact, adaptive  and lightweight representation that probabilistically defines the underlying 3D surface.
We present \ourmethod{}, a neural network that learns to generate {\hgm{}s} which are characteristic of the shape class, and also coincide with the input point cloud.
\ourmethod{} is trained over a collection of shapes to learn a \emph{class-specific prior}.
The hierarchical representation has two main advantages: (i) \emph{coarse-to-fine} learning, which avoids converging to poor local-minima; and (ii) (\emph{an unsupervised}) consistent partitioning of the input shape. We show that as a generative model, \ourmethod{} learns a meaningful latent space which enables generating consistent interpolations between existing shapes, as well as synthesizing novel shapes. We also present a novel framework for rigid registration using \ourmethod{}, that learns to disentangle orientation from structure of an input shape.

\end{abstract}
\section{Introduction}
Point clouds are a common and simple representation of 3D shapes.
They can be directly obtained through scanning devices, which sample the surface of a 3D object.
A major drawback of the point cloud representation is the inherent dependence on the particular sampling pattern, making it sensitive to occlusions, noise and sparse sampling. Moreover, they are \textit{unordered} and \textit{irregular}, and each individual point sample does not carry any non-local information~\cite{bronstein2017geometric, li2018pointcnn, qi2017pointnet,wang2019dynamic}.

The Gaussian mixture model (GMM) is an alternative representation for 3D shapes \cite{ben20183dmfv, eckart2018hgmr}. 
Unlike the point cloud, which uses an arbitrary specific discrete set of samples, GMMs use a fixed and prescribed number of Gaussians that probabilistically define the underlying 3D surface. GMMs are a compact and lightweight representation, which excel in representing a sparse and non-uniform point cloud. They are inherently \textit{adaptive}: representing intricate details with more Gaussians, while large smooth regions can be represented with a smaller number of Gaussians. 

In this work, we present the use of GMMs as an intermediate and compact representation for point cloud processing with a neural network, called \emph{\ourmethod{}}. \ourmethod{} is trained over a set of shapes to learn a \emph{class-specific prior}. For a given point cloud, \ourmethod{} learns a set of Gaussians  
which are characteristic of the shape class, and also coincide with the input point cloud. In other words, \ourmethod{} provides additional geometric information which is otherwise missing from the input point cloud.
Moreover, \ourmethod{} is a \textit{structured} representation, where subsets of Gaussians represent semantic spatial regions of the shape. 
This provides a means to consistently parse a diverse set of shapes, which is learned without any explicit ground-truth labels (Figure~\ref{fig:teaser}).
We demonstrate the advantage of a learned GMM representation for the task of partial point cloud registration and shape generation.

We show that a major challenge in training a neural network to directly partition shapes into distinct regions for each GMM is a loosely defined objective, which is highly susceptible to local minima. To alleviate this issue, we learn in a \textit{coarse-to-fine} manner, through a hierarchical GMM (\hgm{}) framework. We propose a network which learns to subdivide the input points into distinct groups that are modeled by GMMs at different shape scales. This promotes and encourages more effective learning; since the GMMs at the bottom of the hierarchy concentrate on learning smaller spatial regions, while the top-level GMMs learn larger regions.

Although the input point cloud is unordered, \ourmethod{} learns to output GMMs in a consistent and meaningful order. 

This significant property motivates us to use \ourmethod{} as a generative model trained on collection of shapes. We show that the learned latent space enables generating meaningful interpolations between existing shapes, as well as synthesizing novel shapes.
Moreover, we present a novel framework for rigid registration using \ourmethod{}, which learns to disentangle orientation from the input shape.
To facilitate registration of partial shapes, we train the network to receive partial shapes, and generate GMMs as if they were complete.
This enables \ourmethod{} to perform non-trivial registrations which implicitly considers the overlap between missing regions.
We compare to state-of-the-art rigid registration approaches on partial shapes with large rotation angles demonstrating the applicability of \ourmethod{}. 

Finally, another important property of the \hgm{} representation is that it enables a more efficient loss function compared to voxels $\mathcal{O}(n^3)$ or point clouds $\mathcal{O}(n^2)$ which are bound by the resolution of the grid or points, respectively. By contrast, the complexity of a GMM loss function depends on the number ($k$) of Gaussians $\mathcal{O}(n \cdot k)$ ($k \ll n$), and the \hgm{} loss complexity is $\mathcal{O}(n \cdot \log{} k)$ \cite{eckart2016accelerated}.

\section{Related work}
\textbf{Deep networks for irregular 3D shapes.} Pointnet~\cite{qi2017pointnet} first proposed a neural network which operates directly on irregular and unordered point clouds through $1 \times 1$ convolutions followed by global (max) pooling. Several follow-up works proposed incorporating local neighborhood information. For example, Pointnet++~\cite{qi2017pointnetpp} used a hierarchical network to capture local information. PointCNN~\cite{li2018pointcnn} proposed learning a $\chi$-transformation on point clouds, without using global pooling. Wang et al.~\cite{wang2019dynamic} build Euclidean neighborhoods by building a graph from point samples. SPLATNet~\cite{Su_2018_CVPR} represents points in a high-dimensional lattice. Another class of approaches propose irregular convolution and pooling operators on 3D meshes~\cite{hanocka2019meshcnn} or treats shapes as a graph~\cite{monti2017geometric}. For more details on learning on non-Euclidean data (\emph{geometric deep learning}) refer to \cite{bronstein2017geometric}.

\textbf{GMMs for 3D data.} 
Ben-Shabat et al.~\cite{ben20183dmfv} suggested using GMMs to represent 3D shapes, by pre-defining partitions of the point cloud and calculating the associated GMM. The parameters of the GMM are used as input features to a neural network for the task of classification, which was later extended in \cite{ben2019nesti} for normal estimation.
Unlike previous works, \ourmethod{} uses the networks loss function to learn how to best partition the point cloud for the task at hand. Moreover, instead of using the parameters of the GMM as features, the learned GMMs are used directly as the 3D shape representation.

Eckart et al.~\cite{eckart2018hgmr} also propose using a hierarchical GMM representation for the task of shape registration.
 
Yet, their hGMM is generated without learning. Thus, unlike our learned hGMM, it does not include any prior of a training set, nor any implicit information about missing parts.
Moreover,~\cite{eckart2018hgmr} does not provide a latent space as \ourmethod{} does, which enables shape generation.

\textbf{Generative models for 3D shapes.}
In recent years, various works proposed leveraging the power of deep generative models for 3D shape generation. Achlioptas et al.~\cite{achlioptas2017learning} pioneered a deep generative model which directly generated sparse and irregular point clouds. SO-NET~\cite{li2018so} used hierarchical feature structures for generative as well as discriminative tasks. A generative adversarial network for point clouds was proposed in~\cite{li2018point}. Recently, StructureNet~\cite{mo2019structurenet} use graph neural networks to jointly learn geometry and inter-part training on a collection of shapes. SDM-NET proposed generating deformable mesh parts using a VAE \cite{gaosdmnet2019}. Surface Networks~\cite{kostrikov2018surface} propose a generative model for 3D surfaces via a Dirac operator.
Although not a generative model, Tulsiani et al.~\cite{tulsiani2017learning} demonstrated that learning to fit primitives to shapes is an effective approach for exploiting visual similarities in the data.
Our work proposes a novel approach for shape generation via a learned \hgm{} partition. The learned GMMs correspond to consistently segmented regions across a diverse set of shapes, without using any explicit correspondence in the loss function.

\textbf{Shape registration.}
Shape registration is a well-studied problem with a variety of different techniques proposed over the years. There are two main classes of registration approaches: estimating global (rigid) or local (non-rigid) transformations between two potentially partial shapes. A popular technique for rigid registration is to apply RANSAC to find three matching points \cite{fischler1981random, chen1999ransac}. In 4PCS~\cite{aiger20084} a technique for filtering the number of trials required, greatly improves efficiency. Often, rigid registration is followed by non-rigid registration for additional refinement. A popular approach for local registration is ICP~\cite{besl1992method, chen1992object} and its many variants~\cite{rusinkiewicz2001efficient,Segal09Generalized}; which can also be used to compute a rigid transformation. 


Recently, there have been efforts to apply deep neural networks for the task of rigid~\cite{su2015render} and non-rigid~\cite{hanocka2018alignet, Groueix_2019} registration, which can be faster and more robust than classic techniques. 

PointNetLK~\cite{aoki2019pointnetlk} proposed a recurrent PointNet framework for rigid registration, which is more robust to initialization and missing parts compared to ICP.

%
Instead of learning to align one shape to another, we train \ourmethod{} to orient shapes into a canonical pose, thereby indirectly calculating the transformation between two shapes.

\section{Method}
Our solution aims at representing a point cloud by a set of GMMs. To this end, we design an encoder-decoder based framework that generates a set of GMMs for given a point cloud.
The encoder generates a latent vector from the input point cloud, and the decoder reconstructs a GMM from the latent vector. 
To train the network, we propose a maximum-likelihood based loss function that maximizes the probability that the point cloud was generated by the estimated Gaussians.
To improve performance and avoid converging to local minima, we learn a hierarchical GMM (\hgm{}) instead of a single scale GMM. Our method follows \al{Eckart} \cite{eckart2016accelerated}, who demonstrated that \hgm{} is an effective tool for representing $3$D point clouds.

In the following, we describe the \hgm{} representation (Section~\ref{sec:hgmm}) \rh{using the notation} of \cite{eckart2016accelerated}, and the \ourmethod{} architecture (Section~\ref{sec:pointgmm}). Finally, we describe how \ourmethod{} can be used for  shape generation (Section~\ref{sec:generation}) and registration (Section~\ref{sec:registration}).

\subsection{Hierarchical GMM}
\label{sec:hgmm}
\ah{
The \hgm{} can be \rh{viewed} as a tree of Gaussians where the child of each node are \rh{a} refined Gaussian mixture of their parent.
The first level (root) is composed from a mixture of $J$ (overlapping) weighted 3-dimensional multivariate Gaussians $\Theta^{l=1}_{j} = \left\{ \pi_j, \mu_j, \Sigma_j \right\}$.
Given a point cloud $\pc$ of size $N$, its probability of being generated by $\Theta^{l=1}$ is:
\begin{equation}
  p\left(\pc \vert \Theta^{l=1}\right) = \prod_{i=1}^{N} p\left(\pc_{i} \vert \Theta^{l=1}\right) = \prod_{i=1}^{N} \sum_{j=1}^{J} \pi_j  \mathcal{N}\left(\pc_{i} \vert \Theta^{l=1}_{j}\right).
\end{equation}

Based on this probability, we define a set of $C$ latent variables $c_{ij}$ that represent the binary associations between a point $\pc_{i}$ and a Gaussian $\Theta^{l=1}_{j}$ in the mixture model.
We calculate the posterior for all $c_{ij} \in C$ given $\Theta^{l=1}$ by
\begin{equation}
\gamma^{l=1}_{ij} \defeq E\left[c_{ij}\right]= \dfrac{\pi_{j} p\left(\pc_{i} \vert \Theta^{l=1}_{j}\right)}{\sum_{j'=1}^{J} \pi_{j'}  p\left(\pc_{i} \vert \Theta^{l=1}_{j'}\right)}.
\end{equation}

The next levels in the hierarchical model are defined recursively. Given a point cloud $\pc$,
At the root, we calculate the probabilities $p\left(\pc_{i} \vert \Theta^{l=1}_{j}\right)$ and the posteriors $\gamma^{l=1}_{ij}$ for each point and Gaussian in the mixture.
Each $\Theta_{j}^{l=1}$ can then be refined as another Gaussian mixture of size $\hat{J}$:
\begin{equation}
  p\left(\pc \vert \gamma^{l=1}, \Theta^{l=2}\right) = \prod_{i=1}^{N} \sum_{j=1}^{\hat{J}} \pi^{l=2\vert|1}_{j}  p\left(\pc_{i} \vert \Theta^{l=2\vert|1}_{j}\right)
\end{equation}
where the superscript $l=2|1$ indicates the selection of Gaussian parameters at level 2 given the parent node at level 1. Each refinement term is a Gaussian mixture that satisfies $\sum_{j=1}^{\hat{J}} \pi^{l=2\vert|1}_{j} = 1$.
%
In the recursive step, we follow \rh{a \textit{hard partitioning} scheme, where we assign each point in the point cloud to a mixture model. For a point with posteriors $\gamma^{l=d-1}_{ij}$ in level $d - 1$, the point is assigned to the parent with the highest expectation in level $d-1$, \textit{i.e.}, that corresponds to $\argmax{j} \gamma^{l=d-1}_{ij}$.}

\textbf{Log-likelihood.} Finally, given a point cloud $\pc$, \rh{the} log-likelihood for each depth in the hierarchy is given by
\begin{multline}
 \ell_{\hgm{}}\left(\pc\vert \Theta^{l=d} \right) = \\
 \begin{cases} \log{p\left(\pc \vert \Theta^{l=1}\right)} &\mbox{if } d=1  \\ 
    \log{p\left(\pc \vert \gamma^{l=d-1}, \Theta^{l=d}\right)}& \mbox{else}.
 \end{cases}
\end{multline}

\textbf{\hgm{} sampling.}
When sampling from the \hgm{}, we refer to the \rh{leaf} nodes of the tree at level $D$ as a single mixture model to sample from. Therefore, each leaf weight is scaled by its ancestors' weights, \textit{i.e.,} the fixed leaf weight becomes
\begin{equation}
\hat{\pi}_{\ind} = \prod_{d=1}^{D} \pi^{l=d\vert|d-1}_{j}.
 \label{eq:sampling}
\end{equation}
}

\subsection{\ourmethod{}}
\label{sec:pointgmm}

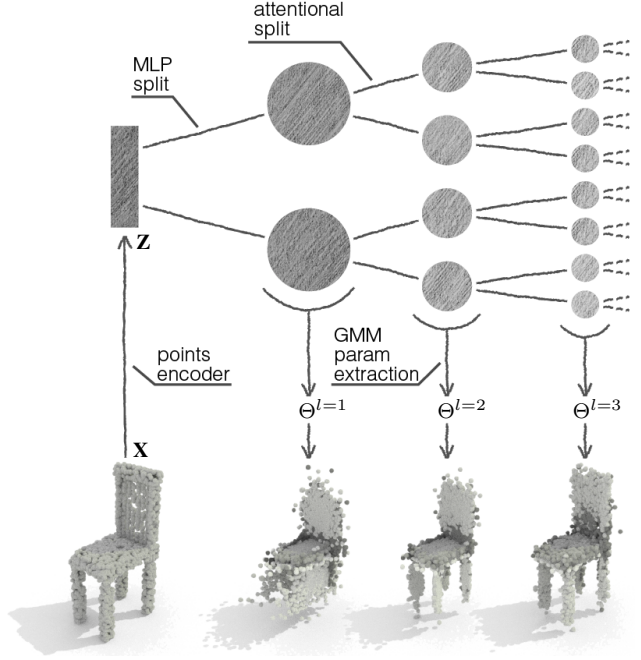
\begin{figure}
    \begin{overpic}[width=\columnwidth]{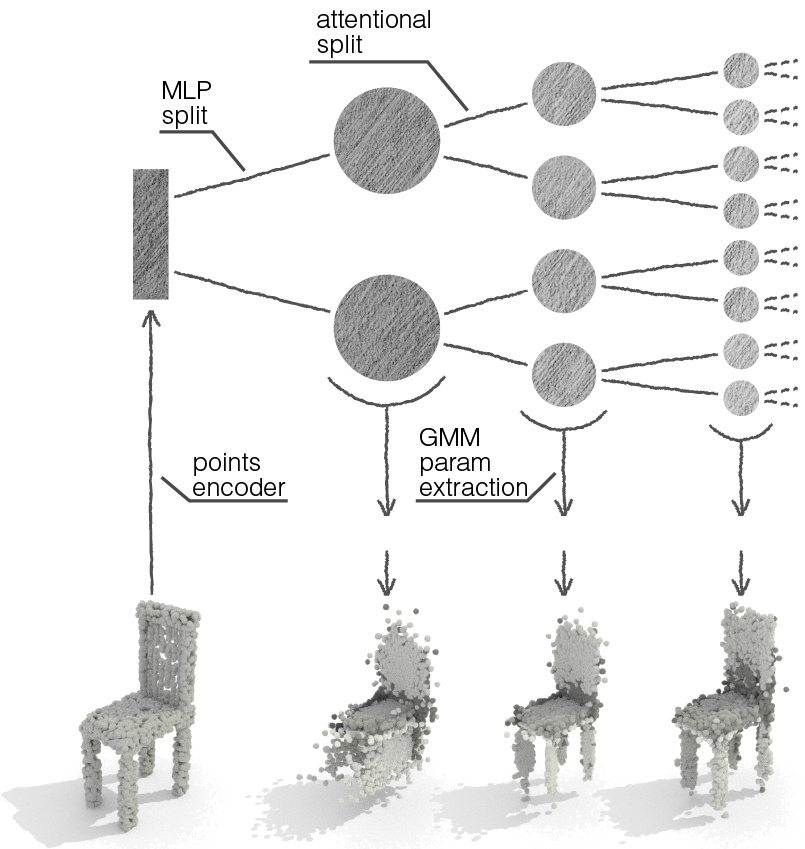}

        \put(19,30){\footnotesize{$\textbf{X}$}}
        \put(19.5,61.5){\footnotesize{$\textbf{Z}$}}
        \put(44,36){\footnotesize{$\ah{\Theta^{l=1}}$}}
        \put(65,36){\footnotesize{$\ah{\Theta^{l=2}}$}}
        \put(85.5,36){\footnotesize{$\ah{\Theta^{l=3}}$}}
    \end{overpic}
    \caption{{\em Method Overview}. \ourmethod{} learns a hierarchical GMM representation of the input $\pc$. Each depth $\depth{}$ of the tree corresponds to a group of GMMs (with parameters $\Theta^\depth{}$) representing the input distribution at different resolutions.}
    \vspace{-0.2cm}
    \label{fig:0301diagram}
\end{figure}

\ourmethod{} proposes a novel encoder-decoder network that builds an \hgm{} tree from an input point cloud. We demonstrate two different use cases of \ourmethod{}: shape generation and registration. The encoder architecture is task-specific, and we use the same \ourmethod{} decoder for both tasks.

The \ourmethod{} decoder network is illustrated in Figure~\ref{fig:0301diagram}. First, an input point cloud $\pc$ goes through an encoder network resulting in a latent vector $Z$. Then, the embedding is used by the decoder to generate the \hgm{} representation. Starting at the root (with latent vector $Z$), the decoder generates the nodes in the \hgm{} tree in a top-down \textit{splitting} process. For each node, the network generates a feature vector, which is used to extract the GMM parameters.

\textbf{MLP split.}
For every depth in the tree, the decoder splits each node into children.
We apply a \ah{multi-layer perceptron (MLP)} on each node. The output of the MLP is a feature map which is used later to extract the GMM parameters. Specifically, the MLP creates a feature vector $h$ \ah{that \rh{represents} each Gaussian in the mixture}. Starting at the root, the initial latent vector is the encoder output $Z$. 

\textbf{Attentional split.}
\ah{After the first MLP split, we have the first level in the \hgm{} that represents a single Gaussian mixture.}
Since we want to \textit{combine} information between siblings\ah{, \rh{\emph{i.e.,}} children of the same parent,} we utilize a self-attention module \cite{vaswani2017attention}. This enables effective information passing between siblings to generate a higher quality mixture. Following the first MLP split into children, each child will split recursively into new children using attentional splits.

To calculate the self-attention, we follow \cite{liu2018generating,vaswani2017attention}, and compute for each node $\ind{}$, its query $\inreals{Q_{\ind{}}}{d_{k}}$, key $\inreals{K_{\ind{}}}{d_{k}}$ and value $\inreals{V_{\ind{}}}{h}$. This is done using three fully connected layers  such that 
$Q_{\ind{}} = f_Q(\ind{})$, $K_{\ind{}} = f_K(\ind{})$, $V_{\ind{}} = f_V(\ind{})$, which share the same parameters for all nodes of the same depth.
The output of these layers are used for calculating the attention weights \ah{between the siblings $Si\left(\ind{}\right)$ of each node $\ind{}$}, which are given by $\hat{\alpha}_{\ind{}} = \left \{ \dfrac{Q_{\hat{j}}^{T}{K_{\hat{j}}}}{\sqrt{d_k}} \vert \ \hat{j} \in Si\left(\ind{}\right) \right \} $. As in regular attention, we transform this vector into a distribution by $\alpha_{\ind{}} = \smax\left({\hat{\alpha}}_{\ind{}}\right)$. The node descriptor before splitting is $\sum_{\hat{j} \in Si(\ind{})} \alpha_{\ind{},\hat{j}}V_{\hat{j}}$, which is splitted into its children by a MLP with one hidden layer as described above. We repeat this process of attention followed by a MLP split until reaching the pre-determined depth.

\begin{figure}[h]
    \centering
    \begin{overpic}[width=0.45\textwidth]{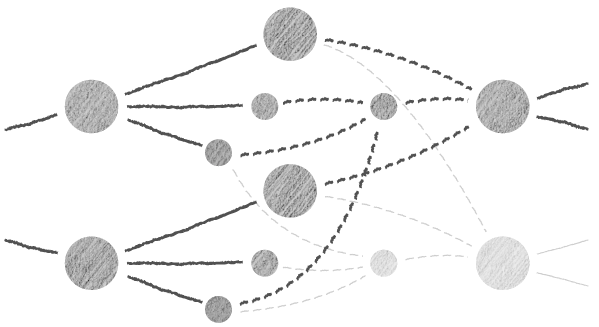}

\put(59,13){\textcolor{grey}{\footnotesize{$\alpha$}}}
\put(59,39){\footnotesize{$\alpha$}}

\begin{turn}{3}
\put(36.5,9.7){\footnotesize{$Q$}}
\put(37.5,36){\footnotesize{$Q$}}
\put(26,36.8){\tiny{$f_{Q}$}}
\put(24.4,10.7){\tiny{$f_{Q}$}}
\end{turn}

\begin{turn}{20}
\put(28.3,11.3){\footnotesize{$V$}}
\put(37.5,36){\footnotesize{$V$}}
\put(22.5,36.5){\tiny{$f_{V}$}}
\put(13,11.9){\tiny{$f_{V}$}}
\end{turn}

\begin{turn}{-20}
\put(10.1,9.9){\footnotesize{$K$}}
\put(1.,34.6){\footnotesize{$K$}}
\put(-5.8,35){\tiny{$f_{K}$}}
\put(3.5,10.4){\tiny{$f_{K}$}}

\end{turn}

\end{overpic}
    \vspace{-0.5cm}
    \label{fig:0302_attention_diagram}
\end{figure}

\textbf{GMM parameter extraction.}
\ah{The children of each node in the \hgm{} tree corresponds to a GMM. }
Each \ah{child node} $\ind{}$ contributes one Gaussian $N(\mu_{\ind{}}, \Sigma_{\ind{}})$ with weight $\pi_{\ind{}}$ to the GMM.
We extract the Gaussian parameters from the feature vector $h$ of each node by applying an MLP with one hidden layer.
%
The output of this MLP $\in \reals^{16}$, corresponds to the parameters $\left\{\inreals{\hat{\pi}_{\ind{}}}{}, \ \inreals{\mu_{\ind{}}}{3}, \ \inreals{\hat{U}_{\ind{}}}{3 \times 3}, \ \inreals{\sqrt{\lambda_{\ind{}}}}{3}\right\}$, which are used afterwards to create the parameters
$\Theta_{\ind{}} = \left\{\pi_{\ind{}}, \mu_{\ind{}}, \Sigma_{\ind{}}\right\}$ of each Gaussian in the GMM.

We ensure that the mixture weights sum to probability $1$ by applying the $\smax$ function to all the node weights $\pi_i$ in the group of siblings.
The covariance $\Sigma_{\ind{}}$ is calculated from the eigen-decomposition $\Sigma_{\ind{}} = U^{-1}_{\ind{}} D_{\ind{}} U_{\ind{}}$ where $D_{\ind{}}$ is diagonal with the vector ${\lambda}_{\ind{}}$ as its diagonal and $U_{\ind{}}$ is a unitary matrix resulting from applying the Gram-Schmidt orthogonalization process on $\hat{U}_{\ind{}}$. 
This decomposition is required since we would like to restrict $\Sigma$ to be a positive definite matrix (PSD). This characteristic is guaranteed as the decomposition is of positive real eigenvalues and their eigenvectors (columns of $U$).

\textbf{\hgm{} loss.}
As mentioned above, the loss function for optimizing \ourmethod{} is, naturally, the negative log-likelihood $\left(\ell_{\hgm{}}\right)$ of a given point cloud \pc{} under the networks output parameters, that is: 
\ah{\begin{equation}
    \mathcal{L}_{\hgm{}}\left( \pc, \Theta \right) = - \frac{1}{\vert \pc \vert}\sum_{d=1}^{D}\ell_{\hgm{}}\left(\pc \vert \Theta^{l=d} \right).
\label{eq:los_hgm}
\end{equation}

$\ell_{\hgm{}}$ are summed over all depths until reaching the the finer level. In \rh{this} way, the network is restricted to represent the shape by 
\rh{each of the GMMs in} each level. This creates a spatial connection between the tree nodes, where children give a finer description of their Gaussian parent.
}

\vspace{20pt}
\subsection{Shape generation} 
\label{sec:generation}
\ourmethod{} can be trained for shape generation using a variational auto encoder (VAE) framework \cite{kingma2013auto}.
In our experiments, we adopt the PointNet architecture \cite{qi2017pointnet} for the encoder. PointNet encodes a given point cloud $\inreals{\pc}{N \times d}$ by 
employing an MLP on each point. 
PointNet maintains the order invariance property by applying a global max-pooling operator over each dimension resulting in an order invariant embedding $\hat{Z}$.

This embedding contains the parameters $Z_{\mu}$ and $Z_{\sigma}$, which are used for generating the latent vector $Z$ = $Z_{\mu} + \epsilon Z_{\sigma}$, where $\epsilon \sim \mathcal{N}\left(0,I\right)$. This makes our model \textit{generative}, enabling a smooth latent space which can randomly be sampled.

The encoding $Z$ is the input to the \ourmethod{} decoder described above, which outputs the \hgm{} parameters. The shape generation loss is composed of $\mathcal{L}_{\hgm{}}$ ~\eqref{eq:los_hgm} and the KL (Kullback- Leibler) divergence with respect to the Normal distribution \ah{\rh{to encourage a continuous latent space}}
\begin{equation}
    \mathcal{L}_{g} =  \mathcal{L}_{\hgm{}}\left(\pc, \Theta \right) + \gamma \mathcal{L}_{KL}\left[\mathcal{N}\left(Z_{\mu},Z_{\sigma}^{2}\right)\Vert \mathcal{N}\left(0,I\right)\right].
\end{equation}

To generate novel shapes, we can sample or interpolate vectors from the learned latent distribution. Latent vectors are decoded by \ourmethod{} to generate the \hgm{} representation. Given the probabilistic representation defined by the \hgm{}, we can sample the underlying surface of the synthesized shape in any desired resolution.

\vspace{15pt}
\subsection{Shape registration} 
 \label{sec:registration}
 \begin{figure}
\begin{overpic}[width=0.5\textwidth,tics=10]{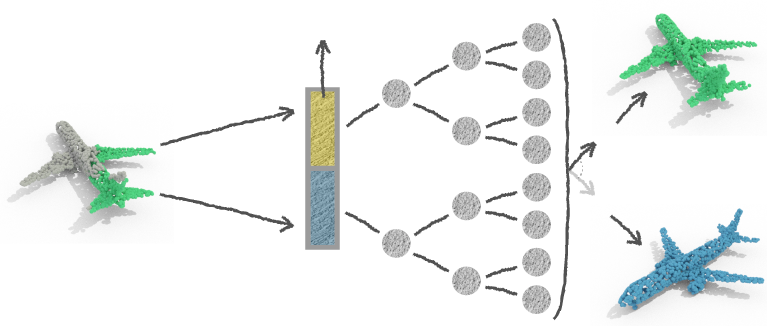}

\begin{turn}{15}
\put(24.7,20.2){\footnotesize{$E_{\tr{}}$}}
\end{turn}

\begin{turn}{-15}
\put(12.2,21.7){\footnotesize{$E_{\sh{}}$}}
\end{turn}

\put(29.5,32.7){\tiny{$MLP$}}
\put(28,38.3){\footnotesize{$\hat{\mathcal{T}}$}}

\put(63.5,25){\tiny{$\Theta_{\tr{}}$}}
\put(64,15.5){\tiny{$\Theta_{\sh{}}$}}

\put(-11,24.5){\tiny{$X$}}

\put(67,36.5){\tiny{$\pcr_{\tr{}}$}}
\put(66,8){\tiny{$\pcr_{\sh{}}$}}

\put(27.4,6.8){\footnotesize{$Z$}}

\put(27.2,14.7){\tiny{$Z_{\sh}$}}
\put(27.2,24.2){\tiny{$Z_{\tr}$}}

\end{overpic}
\caption{\emph{Registration overview.} The input point cloud $\pc{}$ (green) is disentangled into two different embeddings: transformation ($Z_{\tr{}}$) and the shape ($Z_{\sh{}}$); via two parallel encoders $E_{\tr{}}$ and $E_{\sh{}}$. }
\label{fig:reg_diagram}

\end{figure}
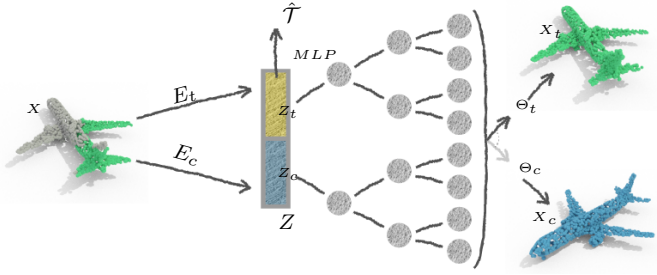

For the registration task, we would like to find a transformation to align two partial point clouds. The core assumption of our approach is that for each shape in our data, there is a canonical form. Thus, our registration strategy is based on finding the transformation of each given shape with respect to its canonical form.


\textbf{Data processing.}
To train our network to learn a canonical position, we use an aligned dataset, ShapeNet~\cite{chang2015shapenet}, for training.
%
We use partially sampled and rotated shapes as an input to our network, generated by the following simple steps. First, we sample the shape surface to get the canonical point cloud $\pcr_{\sh{}}$. Then, the shape is randomly rotated with angle $\phi$ around the $z$ axis, creating $X_{r}$.
The partial point cloud is obtained by sampling a subset of the points. Finally, we translate the point cloud to the center by $v$ resulting in the partial and full point clouds $\pc$ and $\pcr_{\tr{}} = X_{r} + v$.
To summarize, we obtain the input point cloud $\pc$ together with point clouds $\pcr_{\sh{}}$, $\pcr_{\tr{}}$ and transformation $\mathcal{T}=\left\{\phi, v\right\}$, which satisfy $\pcr_{\tr{}} = \mathcal{T} \cdot \pcr_{\sh{}}$. These inputs are used as a supervision to our method as detailed below.


\textbf{Training.}
Our registration framework  (illustrated in Figure~\ref{fig:reg_diagram}) aims to disentangle the shape from the transformation of an input point cloud. First, two different encoders $E_{\tr{}}$ (\textit{transformation}) and $E_{\sh{}}$ (\textit{canonical}) process the input point cloud $\pc{}$ in parallel. The transformation encoder $E_{\tr{}}$ learns to map the Cartesian coordinates of $\pc{}$ to a latent vector $Z_{\tr{}}$ that is \textit{agnostic} to the shape. On the other hand, the shape encoder $E_{\sh{}}$ learns to embed rotation-invariant features~\cite{chen2019clusternet} of $\pc{}$ to a latent vector $Z_{\sh{}}$, which is agnostic to the orientation. The final latent vector $Z$ is a concatenation of both the shape and rotation vectors $Z = Z_{\tr{}} \oplus Z_{\sh{}}$. 

The network learns to disentangle the \textit{shape} and \textit{transformation} of the input point cloud in two different passes. In the \emph{transformation pass}, we estimate the (transformed) complete shape $\pcr_{\tr{}}$ from the (transformed) partial input $\pc{}$. In this scenario the latent vector $Z$ is a concatenation of the output of the transformation encoder and the shape encoder $Z_{\tr{}} \oplus Z_{\sh{}}$. The \ourmethod{} decoder $D$ builds the \hgm{} of the transformed shape from the latent vector $\Theta_{\tr{}} = D\left(Z_{\tr{}} \oplus Z_{\sh{}}\right)$. In addition, the encoding $Z_{\tr{}}$ is passed to an MLP with one hidden layers, which generates an estimate \ah{$\hat{\mathcal{T}}=\left\{\hat{\phi}, \hat{v}\right\}$} of the transformation $\mathcal{T}$.
Therefore, the loss used for the transformation pass is given by
\ah{
\begin{equation}
\mathcal{L}_{\tr{}} = \mathcal{L}_{\hgm{}}\left(\pcr_{\tr{}}, \Theta_{\tr{}}\right) + \mathcal{L}\left(\hat{\mathcal{T}}, \mathcal{T}\right). \\
\label{eq:reg_tr}
\end{equation}
}
The transformation loss \ah{$\mathcal{L}\left(\hat{\mathcal{T}}, \mathcal{T}\right)$} penalizes differences between translation $v$ using the $L_1$-norm and rotation $\phi$ using cosine similarity. Thus, it is defined as
\ah{
\begin{equation}
\mathcal{L}\left(\hat{\mathcal{T}}, \mathcal{T}\right) = \gamma_{1}\mathcal{L}_{1}\left(\hat{v} , v\right) +\gamma_{2}\mathcal{L}_{cos}\left(\hat{\phi}, \phi\right).
\end{equation}
}
In the \textit{shape} pass, we estimate the (canonical) complete shape $\pcr_{\sh{}}$ from the (transformed) partial input $\pc{}$. 
Since the input features of $\pc{}$ to $E_{\sh{}}$ are rotation-invariant, the \textit{shape} latent vector $Z_{\sh{}}$ is agnostic to the orientation.
Since we want the network to disentangle the shape from the orientation, we do not use the transformation encoder, which gets Cartesian coordinates to estimate the transformation, and set the transformation latent vector with zeros $\vec{0}$. This leaves the final latent vector $Z$ to be the output of the shape encoder concatenated with zeros $\vec{0} \oplus Z_{\sh{}}$. The \ourmethod{} decoder $D$ builds the \hgm{} of the transformed shape from the latent vector $\Theta_{\sh{}} = D\left(\vec{0} \oplus Z_{\sh{}}\right)$. Clearly, in the shape pass, there is no MLP-transformation update. Therefore, the loss for the shape pass is given by
\begin{equation}
\mathcal{L}_{\sh{}} = \mathcal{L}_{\hgm{}}\left(\pcr_{\sh{}}, \Theta_{\sh{}}\right).
\end{equation}

Our proposed architecture learns to align shapes into a global canonical orientation. Thus, we can easily obtain the alignment between two shapes by inverting the transformation of one and applying it to the other.

\begin{figure}
\begin{subfigure}[t]{\columnwidth}
\includegraphics[width=\columnwidth]{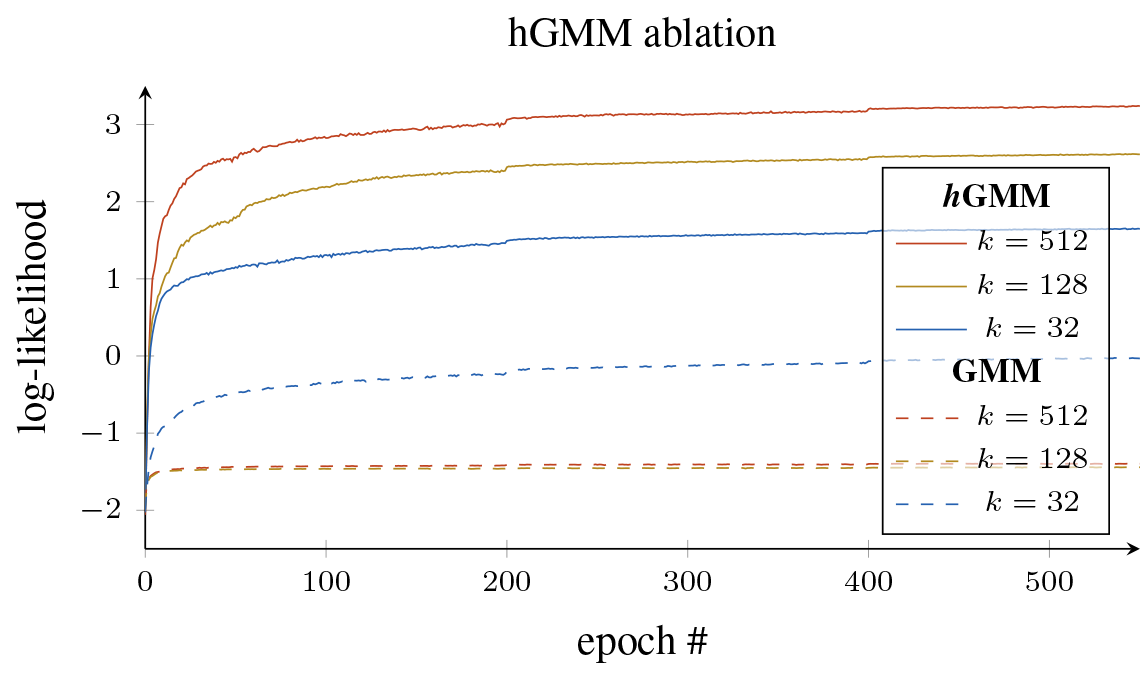}
\phantomcaption
\label{fig:abl_hgm}
\end{subfigure}
\\
\begin{subfigure}[t]{\columnwidth}
\includegraphics[width=\columnwidth]{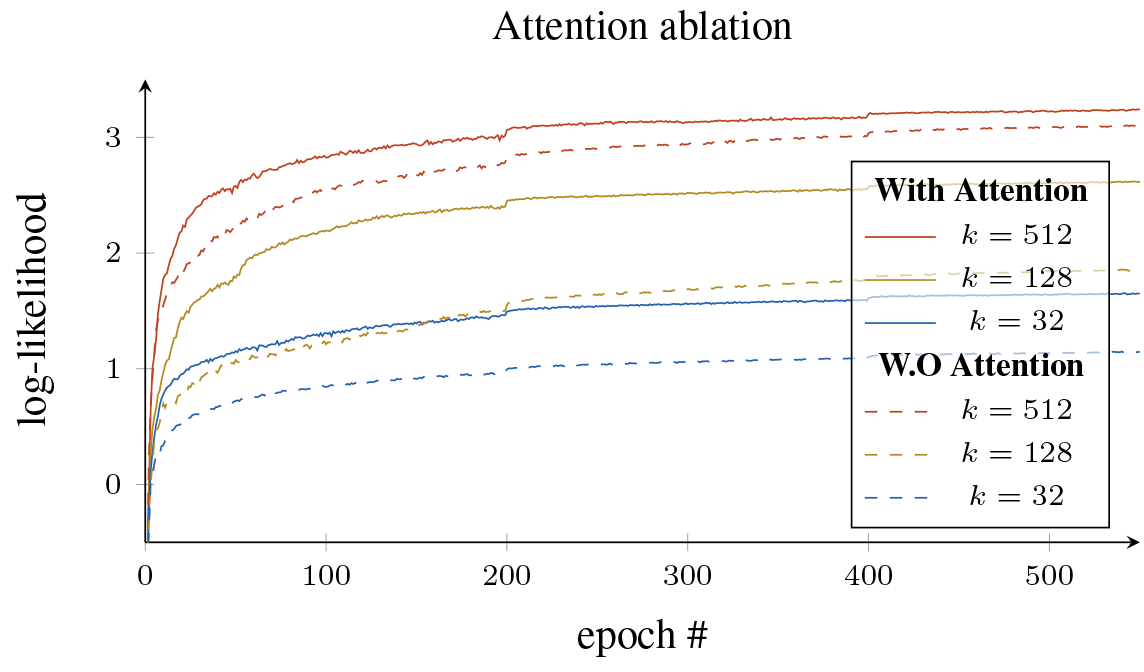}
\phantomcaption
\label{fig:abl_attention}
\end{subfigure}
\caption{\textit{Ablation study.} Top: significance of \hgm{} compared to vanilla GMM. Bottom: impact of using attention.}
\vspace{-.5cm}
\label{fig:abl_graphs}
\end{figure}
\section{Experiments}
\begin{figure*}
    \centering
    \includegraphics[trim={0cm 0cm .7cm 2cm},clip,width=0.35\textwidth]{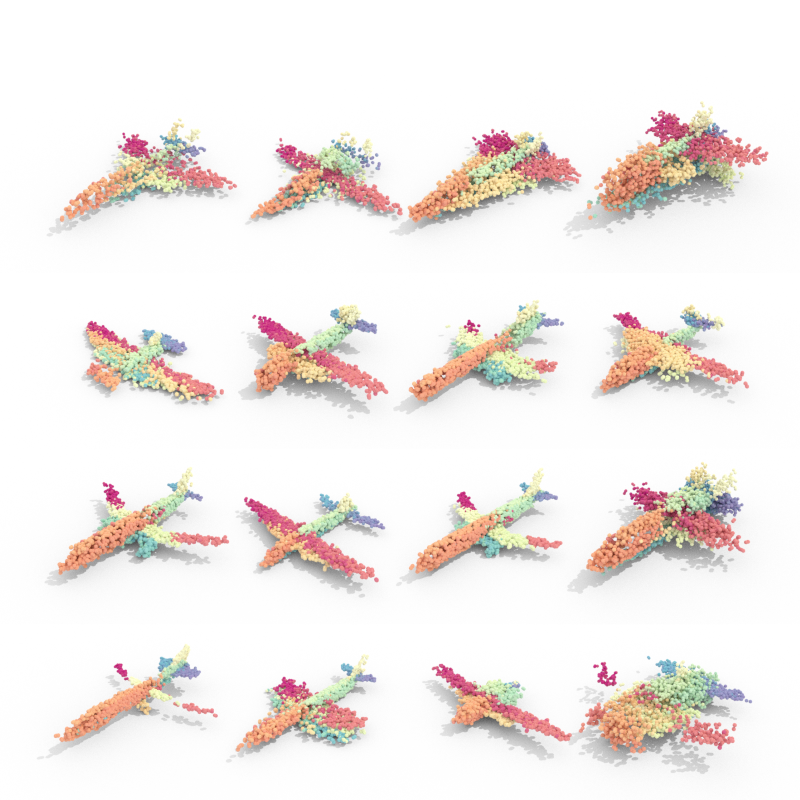}
    \includegraphics[trim={0cm 0cm 1.6cm 0cm},clip,width=0.32\textwidth]{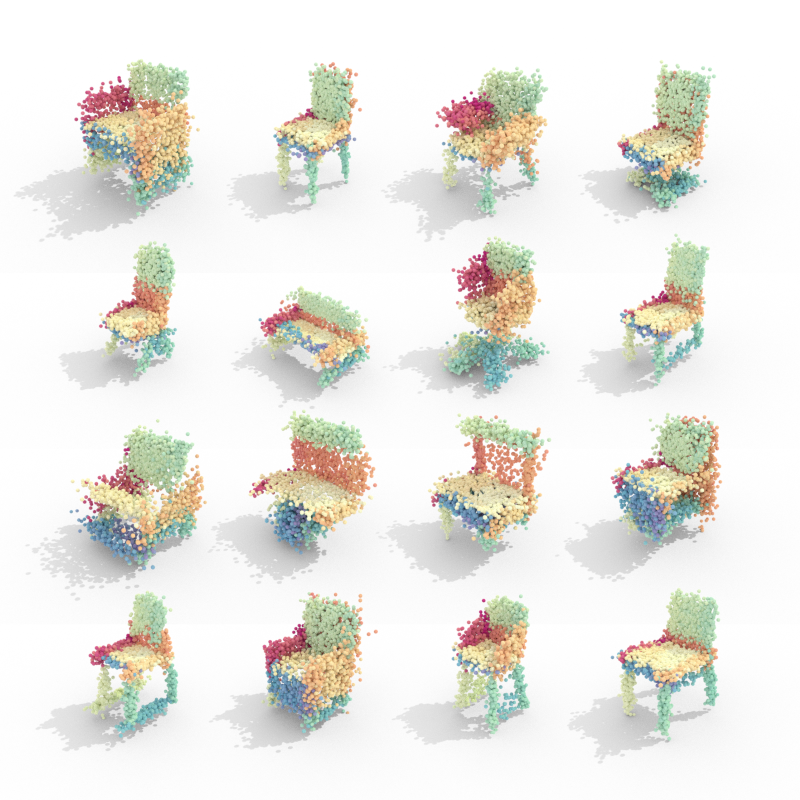}
    \includegraphics[trim={0cm 0cm 1.6cm 0cm},clip,width=0.32\textwidth]{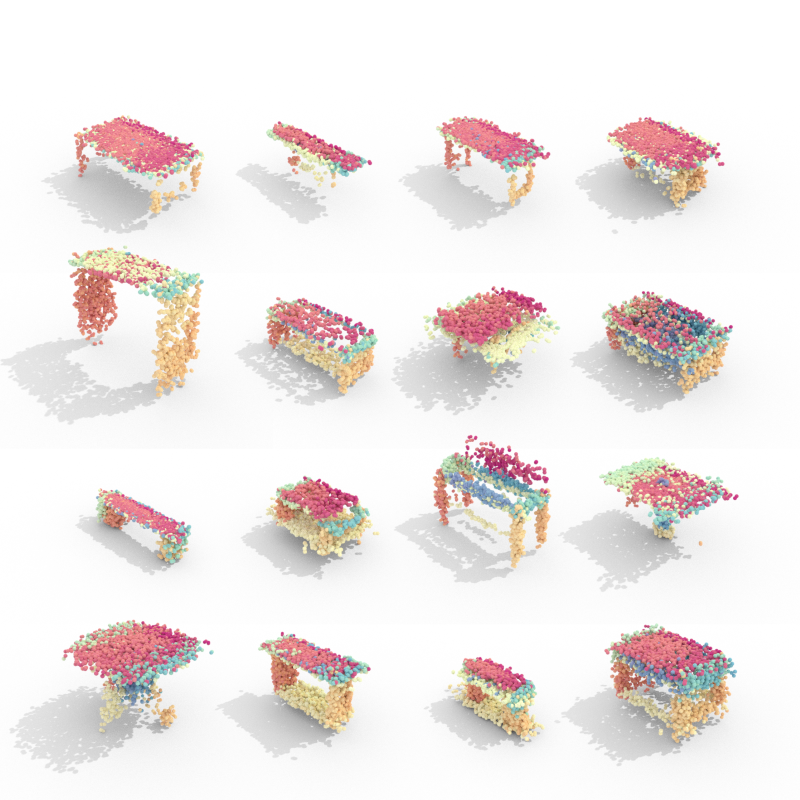}
    \caption{Randomly sampled shapes using a \ourmethod{} generative model.}
    \label{fig:samples}
\end{figure*}

In all experiments we train our network on shapes from the Shapenet dataset \cite{chang2015shapenet}, where each network is trained on a single class.

We train each network for 1000 epochs using Adam optimizer \cite{kingma2014adam} with a learning rate (lr) of $10^{-4}$ and lr decay of 0.5 applied in intervals of 200 epochs.
The value $V_{\ind{}}$, key $K_{\ind{}}$ and query $Q_{\ind{}}$ have dimension $512$, $64$ and $64$ respectively.
All the networks use the following \hgm{} tree form: starting with the first split of the root to $8$ children, with additional $3$ splits to $4$ children each. This results in $8, 32, 128$ and $512$ Gaussian mixtures in each level.

Our PyTorch \cite{paszke2017automatic} implementation as well as the pre-trained models \ah{are available at  \url{https://github.com/amirhertz/pointgmm}.}
\subsection{Shape generation evaluation}
\label{exp:generation}
Following the $VAE$ framework (see Section ~\ref{sec:generation}), we train class-specific networks on the chair, table and airplane datasets. We train with latent vector $Z$ of size $256$ and a weight of $1$ for the $\mathcal{L}_{KL}$ applied with decay of $0.98$ every $100$ epochs.
We use the trained networks to make interpolations between shapes from the dataset (see Figure \ref{fig:teaser}).
In addition, we sample from the latent space to generate new shapes (see Figure \ref{fig:samples}). 
\ah{Quantitative results based on the protocol of \cite{yang2019pointflow} are in the supplementary material, as well as additional qualitative results.}

\textbf{Ablation Study.} 
We perform an ablation study to highlight the contributions of each component of \ourmethod{}. We use the chair dataset on the shape generation task. First, we examine the influence of the hierarchical architecture compared to a vanilla GMM architecture. In the vanilla GMM setting there is no hierarchy, instead the decoder generates all the GMMs once in the last layer of the network. We plot the log-likelihood of the \hgm{} and vanilla GMM nodes vs. epochs in Figure~\ref{fig:abl_hgm} for $k=32$, $128$ and $512$ respectively. Observe the use of \hgm{} is crucial in preventing getting caught in a local minima. Namely, notice that the vanilla GMM with $k=32$ Gaussians was able to learn something, indicating the solution is indeed a coarse-to-fine learning procedure. Moreover, as we increase the number of Gaussians, the \hgm{} network improves, while the vanilla GMM becomes even more susceptible to local minima (performance decreases).
\ah{This phenomenon is also common in \rh{image} GANs, where  a  coarse  to  fine  methodology  was found  useful  to help stabilize training \cite{karras2017progressive}.}

We also verify the importance of the attention module within our network architecture by conducting the 
experiments for the different depths as before while the attention modules are disabled.
We can see in Figure~\ref{fig:abl_attention} that adding attention does improve the log-likelihood during training, especially in the shallower cases (fewer Gaussians). This shows that when the number of Gaussians is limited, it is even more crucial to enable information passing across the \hgm{} nodes using the attention module.

\newcommand{\tfig}{1.9}
\newcolumntype{C}[1]{>{\let\newline\\\arraybackslash\hspace{0pt}}m{#1}}
\begin{figure*}[ht]
    \includegraphics[trim={0cm 1.3cm 0cm 1.3cm},clip,width=\textwidth]{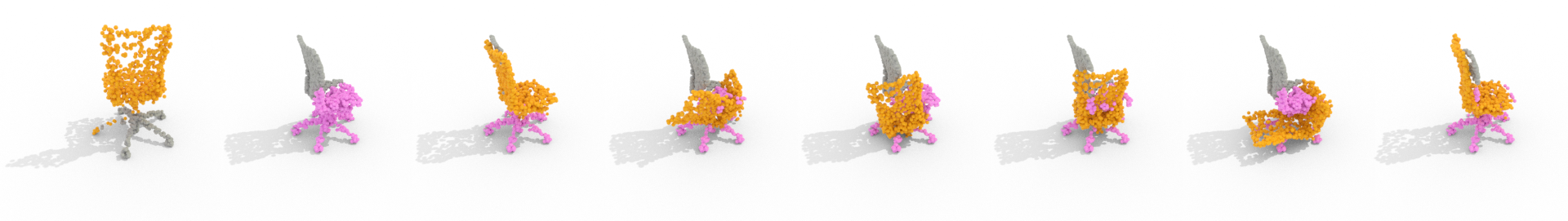}
    \\ 
    \includegraphics[trim={0cm 1.3cm 0cm 2.3cm},clip,width=\textwidth]{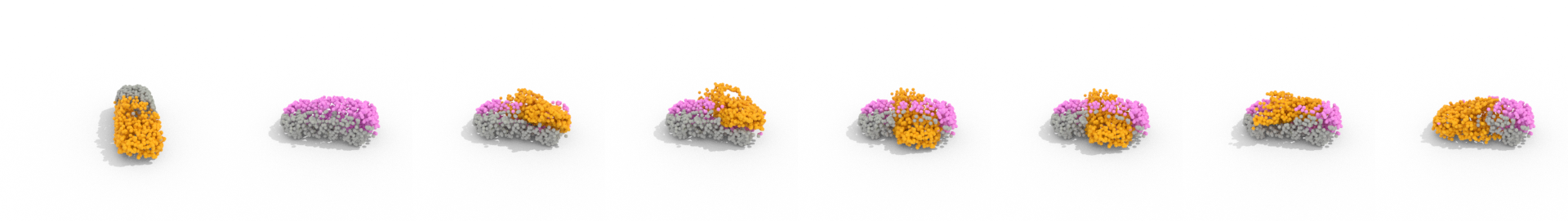}
     \\
     \includegraphics[trim={0cm 1.3cm 0cm 2cm},clip,width=\textwidth]{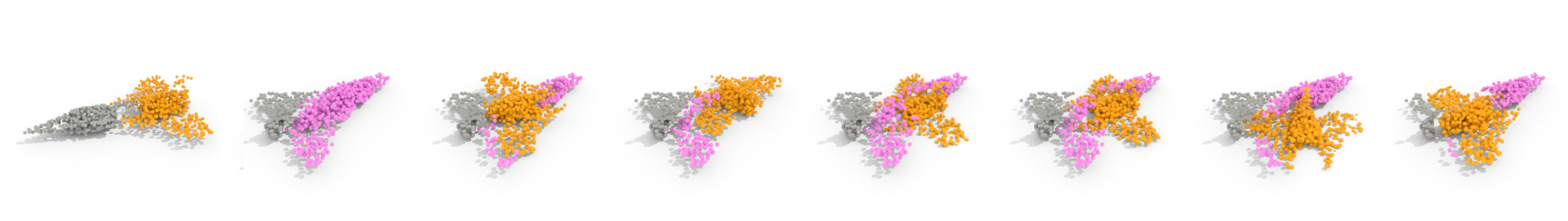}

   \begin{tabular}{C{2.4 cm} C{1.7 cm} C{1.6 cm} C{1.6 cm} C{2. cm} C{1.5 cm} C{1.85 cm} C{1.8 cm}}
   \;\;\;\;\;\; \footnotesize{Source} & \footnotesize{Target} &
    \footnotesize{FPFH \cite{rusu2009fast}} & \footnotesize{S4PCS \cite{mellado2014super}} & \footnotesize{GMMREG \cite{jian2010robust}} & \footnotesize{ICP \cite{besl1992method}} & \footnotesize{PointNetLK \cite{aoki2019pointnetlk}}  & \footnotesize{\ourmethod}
    \end{tabular}
    
    \caption{Qualitative results from the rigid registration comparison. }
    \label{fig:reg_results}
    
\end{figure*}

\setlength{\tabcolsep}{4pt}
\begin{table}[t]\centering

\begin{tabular}{l ? c c c} 
\toprule
shape &
Baseline & Point Decoder & \ourmethod  \\
\midrule
chair  & 0.0612 & 0.1185  & \textbf{0.0382}  \\
car & 0.1514 & 0.2073  & \textbf{0.0447}  \\
airplane & 0.1688 & 0.1817  & \textbf{0.0447}  \\
\bottomrule
\end{tabular} 
\caption{Decoder ablation. The registration results improve when adding a \ourmethod{} decoder (compared to baseline), and excel compared to a vanilla point decoder.}
\label{tab:reg_ablation}
\end{table}

\subsection{Shape registration evaluation}
\label{exp:registration}
\setlength{\tabcolsep}{4pt}
\begin{table*}[t]\centering

\begin{tabular}{l @{\hspace{0pt}} c @{\hspace{0pt}} c ? c c c c  c  c} 
\toprule
shape &
\begin{tabular}{c} max \\ rotation $\left(^{\circ}\right)$ \end{tabular} & 
\begin{tabular}{c}  sampling \\ coverage $\%$ \end{tabular} &

\begin{tabular}{c} RANSAC \\ FPFH \cite{rusu2009fast} \end{tabular} & S4PCS \cite{mellado2014super} & GMMREG \cite{jian2010robust} & ICP \cite{besl1992method} & PointNetLK \cite{aoki2019pointnetlk}  & \begin{tabular}{c} Ours \\ \ourmethod \end{tabular}  \\
\midrule
chair & 30 & 50 - 80 & 0.2113 & 0.3343 & 0.0434 & 0.0430 & 0.1665 & \textbf{0.0226} \\
chair & 30 & 30 - 50 & 0.2804 &  0.3500 & 0.0842 & 0.0824 & 0.2617 & \textbf{0.0496} \\
chair & 180 & 50 - 80 & 0.2481 &  0.3479 & 0.2586 & 0.2578 & 0.2768 & \textbf{0.0232} \\
chair & 180 & 30 - 50 & 0.3132 &  0.3732 & 0.2829 & 0.2817 & 0.3386 & \textbf{0.0574} \\
\midrule
car & 30 & 50 - 80 & 0.1352 &  0.2344 & 0.0399 & 0.04003 &  0.0566  & \textbf{0.0246} \\
car & 30 & 30 - 50 & 0.2134 &  0.2573 & 0.0884 & 0.08774 & 0.1647 & \textbf{0.0552} \\
car & 180 & 50 - 80 & 0.1754 &  0.2411 & 0.2134 & 0.2134 &  0.2288 & \textbf{0.0290} \\
car & 180 & 30 - 50 & 0.2357 &  0.2593 &  0.2354 & 0.2350 & 0.2548 & \textbf{0.0702} \\
\midrule
airplane & 30 & 50 - 80 & 0.0765 &  0.1254 & 0.0632 & 0.0661 &  0.0798 & \textbf{0.0312} \\
airplane & 30 & 30 - 50 & 0.1501 &  0.1637 & 0.1052 & 0.1070 & 0.1301 & \textbf{0.0490} \\
airplane & 180 & 50 - 80 & 0.1485 &  0.1768 & 0.1983 & 0.1979 & 0.2023 & \textbf{0.0350} \\
airplane & 180 & 30 - 50 & 0.1961 & 0.2084 & 0.2293 & 0.2302 & 0.2308 & \textbf{0.0489} \\
     
\bottomrule
\end{tabular} 
\caption{Quantitative comparisons for rigid registration on partial shapes. }
\label{tab:reg_table}
\end{table*}

In the registration experiments, we assume the direction of gravity is known (a common feature of real scanning devices), resulting in partial point clouds in any position or orientation around the $z$ axis.

We test our registration approach (see Section \ref{sec:registration}) by training three class-specific networks on chairs, airplanes and cars. We replace the tables class due to symmetry which makes the registration evaluation ambiguous. We use weights of $20$ and $10$ for the translation $\mathcal{L}_{1}$ and rotation $\mathcal{L}_{cos}$ losses respectively. The dimensions of $Z_{\sh{}}$ and $Z_{\tr{}}$ are $256$ and $128$, respectively.
In all the training scenarios, the input point clouds are uniformly sampled from only $30\%$ to $80\%$ of the underlying shape area and are randomly rotated around the $z$ axis, for example, see \textit{source} and \textit{target} point clouds in Figure~\ref{fig:reg_results}.

\textbf{Evaluation.}
We evaluate the performance of each network on four test sets covering two cases: \textit{medium} ($\leq 30^{\circ}$ ) and \textit{large} rotations ($\leq 180^{\circ}$). We also test different two different ranges of surface area coverage: $50\%$ to $80\%$ and $30\%$ to $50\%$.

Each test consists of 1000 pairs of point clouds, where each pair is sampled from the same shape \ah{and are added with a random Gaussian noise of $\sigma=0.02$}. All test shapes are unseen during training.

Given a pair of source and target point clouds, we compute the rigid transformation from source to target. In order to evaluate alignment accuracy, we compute the mean squared error per point in the transformed source point cloud to the ground-truth target point cloud.
%


\textbf{Comparisons}.
We run the same registration tests on five additional approaches. Two of them, \textit{RANSAC-fast point feature histograms} (FPFH \cite{rusu2009fast}) and \textit{Super 4-points Congruent Sets (S4PCS} \cite{mellado2014super}) are global registration approaches. We also compare to point set registration approaches: the ubiquitous \textit{Iterative Closest Point} (ICP \cite{besl1992method}) and a GMM-based approach \textit{robust point set registration using GMMs} (GMMREG \cite{jian2010robust}). Lastly, we compare to a recent deep learning approach PointNetLK \cite{aoki2019pointnetlk}, which we adapt for our test by training it on the same Shapenet dataset.

The quantitative $MSE$ evaluation results are reported in Table~\ref{tab:reg_table} and qualitative examples in Figure~\ref{fig:reg_results}. Our method achieves out performs the other approaches, most notably when large regions are missing with large rotation angles. 
Observe that the gap between our method and the point set approaches is small in the cases with medium rotations ($\leq 30^{\circ}$ ) and larger sampling coverage ($50\%$ to $80\%$).

\textbf{Ablation study.}
We perform two additional ablation comparisons to demonstrate the utility of \ourmethod{} as a decoder within the registration framework (Figure~\ref{fig:reg_diagram}).

As a \textit{baseline}, we remove the decoder and train only the transformation encoder $E_{\tr{}}$ to output the canonical transformation of an input point cloud.
Thus, this baseline network is trained by the transformation loss $\mathcal{L}_{\mathcal{T}}$ (Equation~\ref{eq:reg_tr}), without the disentanglement \hgm{} losses.

We were also interested in comparing our approach against a simple point-based decoder. Therefore, we replaced \ourmethod{} with an implementation of PointNet auto-encoder \footnote{\url{www.github.com/charlesq34/pointnet-autoencoder}}.  This decoder applies a $4$ layer MLP network on an input vector $Z$ to output a point cloud with $2048$ points. In this setting, the \hgm{} loss is replaced by the dual Chamfer distance.


We ran the decoder ablation on random rotations {($\leq 180^{\circ}$)} with sampling area coverage between $30\%$ to $80\%$ in Table~\ref{tab:reg_ablation}. Observe that using \ourmethod{} performs better than without a decoder at all (baseline). On the other hand, vanilla point decoder did not always do better than the baseline (in fact, worse than the baseline in the chairs set). We believe this is due to the fact that the PointNet decoder struggles to generalize to the ill-posed problem of completing partial-shapes, while \ourmethod{} can define the complete shape with better certainty using a probabilistic model. In other words, the probabilistic \hgm{} framework naturally supports handling partial shapes by predicting large variances in uncertain regions.


\section{Conclusion}

We have introduced a novel framework to transform point clouds into a \hgm{} representation. This representation has various desirable properties, including representing the shape probabilistically and in a coarse to fine manner. This facilitates coping 
with shapes that have missing parts or non-uniform sampling. The coarse levels capture global information of the shape better, while the probabilistic framework may alleviate randomness in the sampling process.


%
The key idea of representing a point cloud using a probabilistic framework was demonstrated to be useful for constructing a generative model for 3D objects,
a problem which is known to be challenging. A common approach for this task directly synthesizes a point cloud, which often leads to a fixed number of points and fixed (uniform) sampling. 
Using \ourmethod{} to learn \hgm{} for shape generation enables sampling the learned distribution to any desired resolution and possibly even non-uniformly.

Note that our model struggles to generate small sharp details (see Figure \ref{fig:samples}). We believe that one source of the problem is the fact that we use a VAE-based model, which is known to produce non-sharp results. Accordingly, incorporating adversarial, or other \textit{detail-sensitive}, losses is expected to improve the representation of finer details.

An intriguing property that was revealed in our representation is that it provides an interpretable and consistent partitioning of a shape class within the \hgm{} leaf nodes (Figure~\ref{fig:teaser}). 
This implies that the learned \hgm{} is structured and captures the shape information. 
We exploited this inherent property to disentangle the shape  from the orientation, and demonstrated how to use this effectively for the task of rigid registration. 


 We believe that \ourmethod{} can be utilized in other shape analysis tasks such as detection, segmentation and reconstruction, to name a few. Note that \hgm{} is a more compact representation compared to the whole point cloud. Thus, such a conversion may allow a more efficient processing in these tasks and serve as an alternative to various sampling methods proposed to improve processing time \cite{dovrat2019learning}.
 
\section*{Acknowledgement}
This work is partially supported by the NSF-BSF grant (No. 2017729) and the European research council (ERC-StG 757497 PI Giryes).

{\small
\bibliographystyle{ieee_fullname}
\bibliography{references}
}

\end{document}